%% file: main.tex
\documentclass{article}

\usepackage{microtype}
\usepackage{graphicx}
\usepackage{subfigure}
\usepackage{booktabs} % for professional tables
\usepackage{xspace}
\usepackage[dvipsnames]{xcolor}

\usepackage{hyperref}
\usepackage{longtable}

\newcommand{\BibTeX}{\rm B\kern-.05em{\sc i\kern-.025em b}\kern-.08em\TeX}

\newcommand{\model}{NES\xspace}

\usepackage[accepted]{icml2023}

\usepackage{amsmath}
\usepackage{amssymb}
\usepackage{mathtools}
\usepackage{amsthm}
\usepackage{enumitem}
\usepackage{subfigure}
\usepackage[capitalize,noabbrev]{cleveref}

\theoremstyle{plain}

\theoremstyle{definition}

\theoremstyle{remark}

\icmltitlerunning{Modeling Dynamic Environments with Scene Graph Memory}

\begin{document}

\twocolumn[
\icmltitle{Modeling Dynamic Environments with Scene Graph Memory}

\icmlsetsymbol{equal}{*}

\begin{icmlauthorlist}
\icmlauthor{Andrey Kurenkov}{stanford}
\icmlauthor{Michael Lingelbach}{stanford}
\icmlauthor{Tanmay Agarwal}{stanford}
\icmlauthor{Emily Jin}{stanford}
\icmlauthor{Chengshu Li}{stanford}
\icmlauthor{Ruohan Zhang}{stanford}
\icmlauthor{Li Fei-Fei}{stanford}
\icmlauthor{Jiajun Wu}{stanford}
\icmlauthor{Silvio Savarese}{salesforce}
\icmlauthor{Roberto Mart{\'i}n-Mart{\'i}n}{utaustin}
\end{icmlauthorlist}

\icmlaffiliation{stanford}{Department of Computer Science, Stanford University}
\icmlaffiliation{salesforce}{Salesforce AI Research}
\icmlaffiliation{utaustin}{Department of Computer Science, University of Texas at Austin}

\icmlcorrespondingauthor{Andrey Kurenkov}{andreyk@stanford.edu}

\icmlkeywords{Machine Learning, Graph Neural Net, Link Prediction, Object Search}

\vskip 0.3in
]

\printAffiliationsAndNotice{}  % leave blank if no need to mention equal contribution

\begin{abstract}
Embodied AI agents that search for objects in large environments such as households often need to make efficient decisions by predicting object locations based on partial information. We pose this as a new type of link prediction problem: \textbf{link prediction on partially observable dynamic graphs}. Our graph is a representation of a scene in which rooms and objects are nodes, and their relationships are encoded in the edges; only parts of the changing graph are known to the agent at each timestep. This partial observability poses a challenge to existing link prediction approaches, which we address. We propose a novel state representation -- Scene Graph Memory (SGM) -- with captures the agent’s accumulated set of observations, as well as a neural net architecture called a Node Edge Predictor (NEP) that extracts information from the SGM to search efficiently.
We evaluate our method in the Dynamic House Simulator, a new benchmark that creates diverse dynamic graphs following the semantic patterns typically seen at homes, and show that NEP can be trained to predict the locations of objects in a variety of environments with diverse object movement dynamics, outperforming baselines both in terms of new scene adaptability and overall accuracy. The codebase and more can be found  \href{www.scenegraphmemory.com}{this URL}.
\vspace{-0.35cm}
\end{abstract}

\input{1-intro}

\input{2-related}
\input{3-problem}

\input{4-method}
\input{5-exp}
\input{6-results}
\input{7-discuss}

\paragraph{Acknowledgments.}
This work is in part supported by Stanford Institute for Human-Centered Artificial Intelligence (HAI), NSF RI \#2211258, ONR MURI N00014-22-1-2740, AFOSR YIP FA9550-23-1-0127, Analog Devices, JPMorgan Chase, Meta, and Salesforce.

\bibliography{ref}
\bibliographystyle{icml2023}

%%%%%%%%%%%%%%%%%%%%%%%%%%%%%%%%%%%%%%%%%%%%%%%%%%%%%%%%%%%%%%%%%%%%%%%%%%%%%%%
%%%%%%%%%%%%%%%%%%%%%%%%%%%%%%%%%%%%%%%%%%%%%%%%%%%%%%%%%%%%%%%%%%%%%%%%%%%%%%%
% APPENDIX
%%%%%%%%%%%%%%%%%%%%%%%%%%%%%%%%%%%%%%%%%%%%%%%%%%%%%%%%%%%%%%%%%%%%%%%%%%%%%%%
%%%%%%%%%%%%%%%%%%%%%%%%%%%%%%%%%%%%%%%%%%%%%%%%%%%%%%%%%%%%%%%%%%%%%%%%%%%%%%%
\newpage
\appendix
\onecolumn
\input{8-appendix.tex}

%%%%%%%%%%%%%%%%%%%%%%%%%%%%%%%%%%%%%%%%%%%%%%%%%%%%%%%%%%%%%%%%%%%%%%%%%%%%%%%
%%%%%%%%%%%%%%%%%%%%%%%%%%%%%%%%%%%%%%%%%%%%%%%%%%%%%%%%%%%%%%%%%%%%%%%%%%%%%%%

\end{document}

%% file: 1-intro.tex
\begin{figure*}
    \includegraphics[width=0.95\linewidth]{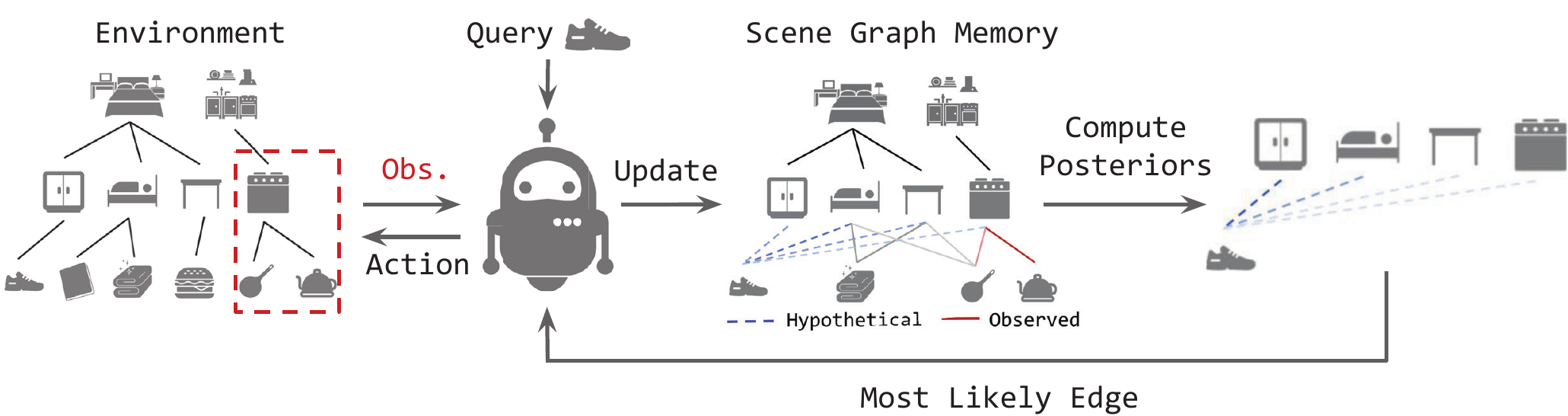}
    \caption{Our problem setup and proposed method: an agent is queried to find target objects in an unknown dynamic household environment where objects change location over time and objects may be added or removed -- a specific instance of the general problem of link prediction in partially observable dynamic graphs. The agent can observe subsets of the true scene graph, which are aggregated in a Scene Graph Memory which is then passed into the Node Edge Predictor model that produces observation-conditioned posterior probabilities of where the query object is. Finally, the agent uses the posterior probabilities to decide on its next action.}
    \vspace{-0.1cm}
    \label{fig:main}
\end{figure*}

\section{Introduction}

Temporal link prediction is the problem of estimating the likelihood of edges being present in the future in a dynamically changing graph based on past observed instances of the full graph~\cite{divakaran2020temporal}. This type of problem appears when analyzing social networks, communication networks, or even biological networks. We investigate a novel instance of this problem: temporal link prediction with partial observability, i.e. when the past observations of the graph contain only parts of it. This setting maps naturally to a common problem in embodied AI: using past sensor observations to predict the state of a dynamic environment represented by a graph. Graphs are used frequently as the state representation of large scenes in the form of \textit{scene graphs}~\cite{johnson2015image,armeni20193d,ravichandran2021hierarchical,hughes2022hydra}, a relational object-centric representation where nodes are objects or rooms, and edges encode relationships such as \texttt{inside} or \texttt{onTop}.
Link prediction could be applied to partially observed, dynamic scene graphs to infer relationships between pairs of objects enabling various downstream decision-making tasks for which scene graphs have been shown to be useful such as navigation~\cite{amiri2022reasoning,santos2022deep},  manipulation~\cite{agia2022taskography,zhu2021hierarchical} and object search~\cite{ravichandran2021hierarchical, xu2022learning}.

In this paper, we study link prediction in dynamic, partially observable graphs with a focus on using this formalism to perform \textbf{object search} with an embodied AI agent in a large scene. Although this is a popular problem, most past works assumes a static scene with the agent having no prior memory of it, whereas we focus on dynamic scenes with a continually learning agent. 
For that, we first  \textbf{propose a novel state representation} named a \textbf{scene graph memory (SGM)} that encodes the nodes and edges the agent has observed---including those that may no longer be true---in a single graph with reference to the time the nodes and edges were lastly observed. The SGM enables the agent to gradually build up a representation of all of its observations, which can then be the input to a link prediction model.

Existing solutions for dynamic link prediction assume full observability (knowing all the nodes and edges in graphs from prior timesteps) and are not well suited to this new kind of link prediction problem. Therefore, we \textbf{introduce a novel solution for dynamic link prediction in partial observable settings} based on a neural net architecture we call a \textbf{Node Edge Predictor (NEP)}. The NEP is designed to predict the likelihood of a single node's edges with a self-attention mechanism that ``compares'' them.

Lastly, we \textbf{address the lack of existing benchmarks for link prediction with partially observable graphs} by introducing the \textbf{Dynamic House Simulator}. Existing simulators for evaluating embodied agents in household environments \cite{li2021igibson, shen2021igibson, savva2019habitat, gan2020threedworld, deitke2022procthor} change only by actions of the embodied AI agent, whereas the Dynamic House Simulator enables sampling diverse dynamic scenes that realistically evolve over time.

In summary, the contribution of our work is three-fold:
\begin{itemize}[leftmargin=5.5mm]
    \item We propose a new state representation, the Scene Graph Memory, to enable embodied agents to aggregate their observations of a dynamic environment over time.
    \item We introduce a novel neural net architecture, the Node Edge Predictor, and show it can be used to perform link prediction learning on dynamic, partially observable graphs. To the best of our knowledge, this is the first solution to this type of problem.
    \item We present the Dynamic House Simulator, a novel benchmark to evaluate embodied agent performance in dynamic, partially observable environments, and show that our proposed method significantly outperforms multiple baselines.
\end{itemize}

%% file: 2-related.tex
\section{Related Work}

\textbf{Temporal link prediction:}
Link prediction is the problem of predicting the likelihood of unknown links (edges) in graphs. While the community has studied the problem extensively in static graphs (we refer to \citet{kumar2020link} for a comprehensive review), link prediction in dynamic graphs, or temporal link prediction as it is also known, is also an important emerging problem~\cite{divakaran2020temporal}. 
Recently, there has been an increasing amount of approaches based on graph neural networks (GNNs) for non-temporal \cite{zheng2021cold,opolka2022bayesian,zhao2022learning,atzeni2021modeling} as well as temporal link prediction \cite{qu2020continuous,lei2019gcn,skarding2022robust,singh2021edge} due to the capabilities of GNNs to encode and efficiently propagate  graph information for inference. Several works also combine GNNs with transformers ~\cite{yang2021graphformers,jin2022heterformer}, which we also do with our Node Edge Predictor.

We focus on a modified version of the temporal link prediction problem. Motivated by challenges embodied AI agents typically face, we add the property of partial observability: only subsets of the prior input graphs are known. This makes the problem significantly more challenging, as the model must make its predictions based on less information. Furthermore, our problem formulation allows for not just the edges but also the nodes in the graph to be dynamic, which has rarely been the case in prior work~\cite{haghani2017temporal,ran2022predicting}. To our knowledge, this is the first work to introduce the problem of temporal link prediction on partially observable dynamic graphs.

\textbf{Modeling object relations in embodied AI research:}
Often in embodied AI tasks, the relations between objects in the environment provide critical information for decision-making. Previous works have designed various data structures or representations that embed co-occurrence statistics between objects such as probabilistic semantic maps \cite{li2012indoor,veiga2016efficient}, graphical models \cite{kim2019active,kollar2009utilizing,aydemir2013active,lorbach2014prior}, hierarchical models \cite{pronobis2017deep}, extended POMDPs \cite{zheng2022towards}, and scene graphs \cite{kurenkov2021semantic}.

Although object co-occurrence statistics can be learned by the agent from scratch in an environment, it is beneficial to extract such statistics prior to agent deployment from knowledge sources that capture commonsense about object relations, such as hand-coded priors~\cite{lorbach2014prior}, online text~\cite{zhou2012web} or image~\cite{kollar2009utilizing} datasets, or curated knowledge base~\cite{toro2014probabilistic}. We introduce a novel way to extract a scene graph specific co-occurrence statistics through counting of observed object relations in the simulated environments of iGibson and ProcThor~\cite{li2021igibson,deitke2022procthor}.

\textbf{Modeling object locations in scenes:}
As summarized in~\citet{crespo2020semantic}, graphs have commonly been used in the embodied AI literature as memory for tasks that require semantic information. We highlight several recent works that are particularly related to ours. \citet{wu2019bayesian} also uses observations to build a Bayesian probabilistic relation graph of the room connectivity in novel houses, but is focused on static environments with no objects, and does not used link prediction. \citet{kurenkov2021semantic} also uses link prediction on a scene graph for finding objects, but does so in the context of static scenes without the need to learn object dynamics. \citet{dulearning} also attempts to capture the dynamics of object movement by maintaining an object-based memory and training attention-based neural networks, but focuses on visual rather than semantic information and within a much smaller number of locations and objects than us. \citet{rudra2022contextual} also produces probabilities over object locations but does so with a contextual bandits approach over vantage points with no use explicit memory, and for just three comparatively simple environments.  Lastly and most related, the concurrent work of \citet{patel2022proactive} also learns a predictive model of object dynamics with the use of scene graphs and GNNs as well as a household simulator, but assumes full knowledge of past states of the environment rather than dealing with partial observability, does not use the graph as a form of memory, does not make use of linguistic cues for generalization, and focuses on realistically simulating five households rather than procedural generation of any number of households.

%% file: 3-problem.tex
\section{Problem Formulation}
\label{sec:problem}
% link prediction problem for dynamic graphs 
The standard temporal link prediction problem for dynamic graphs is defined as follows~\citep{liben2003link,divakaran2020temporal}:
The state of a dynamic graph at time $t$ is represented as $G_t = (V, E_t)$, where $E_t$ is the set of edges present at time $t$ and $V$ are the nodes. Given past observations of the state of $G$ from time step $0$ to $t$, $G_0,...,G_t$, the goal is to predict the presence of all future edges in the next step, $E_{t+1}$.

% link prediction problem for dynamic graphs with partial observability
Our problem formulation (see Fig.~\ref{fig:main}) is a derivation of the above with three major differences. First, the set of nodes in the graph can change between timesteps, so that $V_t\neq V_{t+1}$. Second, we assume partial observability; instead of having full access to past graph states, $G_0,...,G_t$, our agent has only access to partial observations of these states, $O_0,...,O_t$. Each partial observation, $O_t = (V^O_t, E^O_t)$, contains the current state of a subset of the nodes, $V^O_t \subseteq V$, and the edges, $E^O_t \subseteq E_t$. The content of an observation, $O$, is task-dependent; in our embodied AI context it will be the result of agent actions, representing a realistic sensing operation. 

The third difference is in the scope of the prediction: instead of attempting at estimating the state of \textit{all} edges, our goal is to predict the state of \textit{a subset} of the edges, $E^Q_t \subseteq E_{t+1}$. We assume this subset is associated with query nodes $V^Q_t$ of interest for a downstream task. The query edges as well as the query node may or may not have been seen in a past observation and may or may not change from $G_t$ to $G_{t+1}$. This is a natural setup in several embodied AI tasks such as object search, where the agent does not need to infer the state of the entire environment but only the relevant information to find a specific target objects.

% To summarize, this problem involves receiving and producing a subset of information compared to standard dynamic link prediction, in which $E^O_t = E_t$ and $E_q = E_{t+1}$.

% Household setup
\textbf{Object search with temporal link prediction in dynamic, partially observable graphs:} While our method is applicable to temporal link prediction with any form of graph information, we focus on the problem of reasoning about scene graphs in household environments. In this domain, the nodes in a given scene graph are hierarchically organized: at the top are the room nodes, then furniture nodes, and then object nodes. The edges represent the kinematic relations between nodes, such as \texttt{inside} and \texttt{onTop}. The edges between furniture nodes and object nodes are dynamic, and the underlying environment dynamics modify the edges according to some unknown probability distribution. In other words, objects ``move'' over time as they do in real household settings populated by humans. 
%These modifications result in a sequence of scene graphs $SG_0,...,SG_t$ representing the state of the environment. 

Given this household context, the observations $O_0,...,O_t$ consist of one or more furniture nodes along with the objects connected to these furniture nodes with an edge. The goal of our agent is to correctly predict the furniture node that is connected to a target object node. The agent can choose furniture nodes to receive an observation from, which represents it exploring to observe where objects are. The agent's observations are noisy: some objects may not be observed even if connected to the chosen furniture node. This simulates  realistic perception that may fail to detect objects due to occlusions or other factors. 
%The contents of observations depend on the task and agent actions and will be discussed in sec. \ref{s:es}.

%% file: 4-method.tex
\section{Method}
\label{sec:method}

To study and address the dynamic link prediction problem with partial observability, our work includes several components: (1) The Dynamic House Simulator -- a mechanism to simulate household environments with diverse object layouts and distributions, (2) Scene Graph Memory (SGM) -- a state representation to aggregate agent observations and serve as a basis for learning to predict where objects are, and (3) Node Edge Predictor (NEP) -- a novel neural network architecture that is best suited for predicting the presence of query edges in the scene graph.

\subsection{Dynamic House Simulator}
\label{ss:dhs}
\begin{algorithm}
\caption{Dynamic House Simulator Algorithm}\label{alg:dhs}
\begin{algorithmic}
\REQUIRE Initial prior probability graph $P^{prior}$
\FOR{$i=0,1,\dots,M$}
% \STATE Initialize a new environment instance $Env^i=\{ \}$
\STATE Create a noisy copy of $P^{prior}$: $P^i \leftarrow P^{prior} + N^i_{class}$  
\STATE Sample object instances from $P^i$ and create scene graph $SG^i_0$ at its initial state ($t=0$)
\STATE Create environment dynamics: $T^i \leftarrow P^i + N^i_{instance}$
\STATE Define a new env instance: $Env^i = \{SG^i_0, T^i\}$
\ENDFOR
\FOR{$t=1,2, \dots, t_{end}$}
\FOR{$i=0,1, \dots, M$}
\STATE Evolve scene graph: $SG^i_t \leftarrow T^i(SG^i_{t-1})$ 
\ENDFOR
\ENDFOR
\end{algorithmic}
\end{algorithm}

Due to there not being a well suited benchmark for our task, we created the Dynamic House Simulator to benchmark link prediction in dynamic, partially observable graphs in the context of object search in households.
There exist several simulation frameworks for embodied AI in household environments~\cite{li2021igibson, shen2021igibson, savva2019habitat, gan2020threedworld,deitke2022procthor} but none of them simulate the object movement that results from other agents such as humans interacting and changing object locations. 
Our simulator supports sampling a wide variety of household environments with a distinct set of initial objects, locations, and patterns of object movement over time. 
%This represents realistic object dynamics in houses populated by humans.%, all of which are meant to be similar to what is typically found in real household environments. %Because our focus is on this semantic realism, 

\textbf{Priors graph.} The first component of the simulator is a graph encoding the probabilities of room-furniture and furniture-object relationships for all households (Fig \ref{fig:dhs}). We call this our prior probabilities graph $P^{prior}$ because it represents the commonsense knowledge about any given household environment prior to observing it. The probability of an object being \texttt{inside} or \texttt{onTop} a piece of furniture depends on which room the furniture is in. We compute these probabilities via simple counting of the presence of relationships in different environments of iGibson 2.0~\cite{li2021igibson} and ProcTHOR-10k~\cite{deitke2022procthor}, two simulators with realistic object placements in house environments, but the prior could come from any source of furniture-object-room distribution such as a large language model. 
%Since ProcTHOR-10k has a larger number of object types and more sophisticated procedural generation logic, we default to using probabilities derived from it for any relationships found in both iGibson 2.0 and ProcTHOR-10k\todo{this sentence is unclear. do we use iG2 or not?}. 

%See Appendix TODO for more details. 

%There are two versions of this graph: the \textit{coarse} graph which contains F-O probabilities that are independent of R-F probabilities, and the \textit{detailed} graph which contains F-O probabilities that are dependent on R-F probabilities. To be specific, in the latter case, the probability of an object being inside a furniture depends on which room the furniture is in. %or instance, in the coarse graph a jar has a $50\%$ probability of being inside any shelf and a $50\%$ probability of being inside any cabinet, whereas in the detailed graph a jar has a $50\%$ probability of being inside a shelf in the kitchen but only a $5\%$ probability of being inside a shelf in the bedroom. 

After obtaining the prior probabilities, we manually annotate each node in $P^{prior}$ with the following attributes: a set of adjectives that could describe the object or piece of furniture (e.g. ``blue'', ``metal'', etc.), the min/max number of the object in the environment, and how likely the object is to move between furniture or to disappear from or appear in the environment over time. These attributes enable more varied and realistic simulation of the household environment. Additional details can be found in the Appendix (Sec.~\ref{sa:dhs}).
% For instance, ``apple'' is similar to ``orange'' and ``banana''. For each similar type, the node associated with it is duplicated and added to the graph with its original type replaced with the similar type. The goal is to make sure there are sufficient nodes that have similar patterns of location and movement which the agent can observe.

% Following the additions of each node, the graphs contain 4 unique edge types, 26 unique furniture types, and 100 unique object types, which results in a total of 621 edges with relation probabilities (which correspond to relations observed in the iGibson 2.0 and ProcTHOR-10k data).

\textbf{Scene sampling and evolving.} The priors probabilities, $P^{prior}$, are the basis for the simulator algorithm to sample diverse dynamic environments in the form of scene graphs, which we now describe (Alg.~\ref{alg:dhs} and Fig.~\ref{fig:dhs}). To sample an environment instance, $Env^{i}$, we first inject class-level noise, $N_{class}^{i}$, to the prior graph to obtain the environment-specific relation probabilities, $P^i$. The class-level noise is a function of object class, furniture class, and relationship type. For example, for the specific environment $Env^{i}$, a jar has an $80\%$ probability of being inside a shelf and a $20\%$ probability of being inside a cabinet. The specifics of noise generation are included in the Appendix (Sec.~\ref{sa:dhs}). %This can be done either with the coarse or detailed priors.

Then, the initial scene graph $SG_{0}^{i}$ will be sampled based on $P^i$ in a procedural manner. Given the minimum and maximum bounds, % on the number of rooms in the household, the number of pieces of furniture in each room, and the number of objects inside or on top of each piece of furniture, 
$P^i$ is used to first sample a set of rooms, then a set of furniture items within each room, and then objects for each piece of furniture. Note that there might be multiple instances of the same object type in $SG_{0}^{i}$. To capture the fact that different instances of the same object type are distinct in their appearance or material, each sampled node is associated with a description, in addition to its type. The description is generated by randomly sampling a subset of adjectives from the prior graph. For instance, a ``mug'' node may get the description ``large red mug''.

After this, we create the environment-specific object relation change probabilities, or dynamics for short, $T^{i}$ by injecting instance-level noise $N_{instance}^{i}$ into $P^i$. The instance-level noise $N_{instance}^{i}$ is a function of object description, furniture description, and relationship type. For example, a ``large red mug'' has a $50\%, 30\%, 5\%, 15\%$ chance of being inside of shelf 1, shelf 2, cabinet 1, and cabinet 2 respectively.

At each timestep, the dynamics $T^{i}$ will be used to modify the scene by changing the furniture-object relations in the scene graph $SG_{t}^{i}$ to create $SG_{t+1}^{i}$. $T^{i}$ can also potentially add or remove an object instance based on $P^{i}$. As a result, each environment is defined in terms of its unique initial scene graph and unique object dynamics $Env^{i}=\{SG_{0}^{i}, T^{i}\}$, which conceptually corresponds to a unique household. 

\begin{figure}
    \centering
    \includegraphics[width=0.9\linewidth]{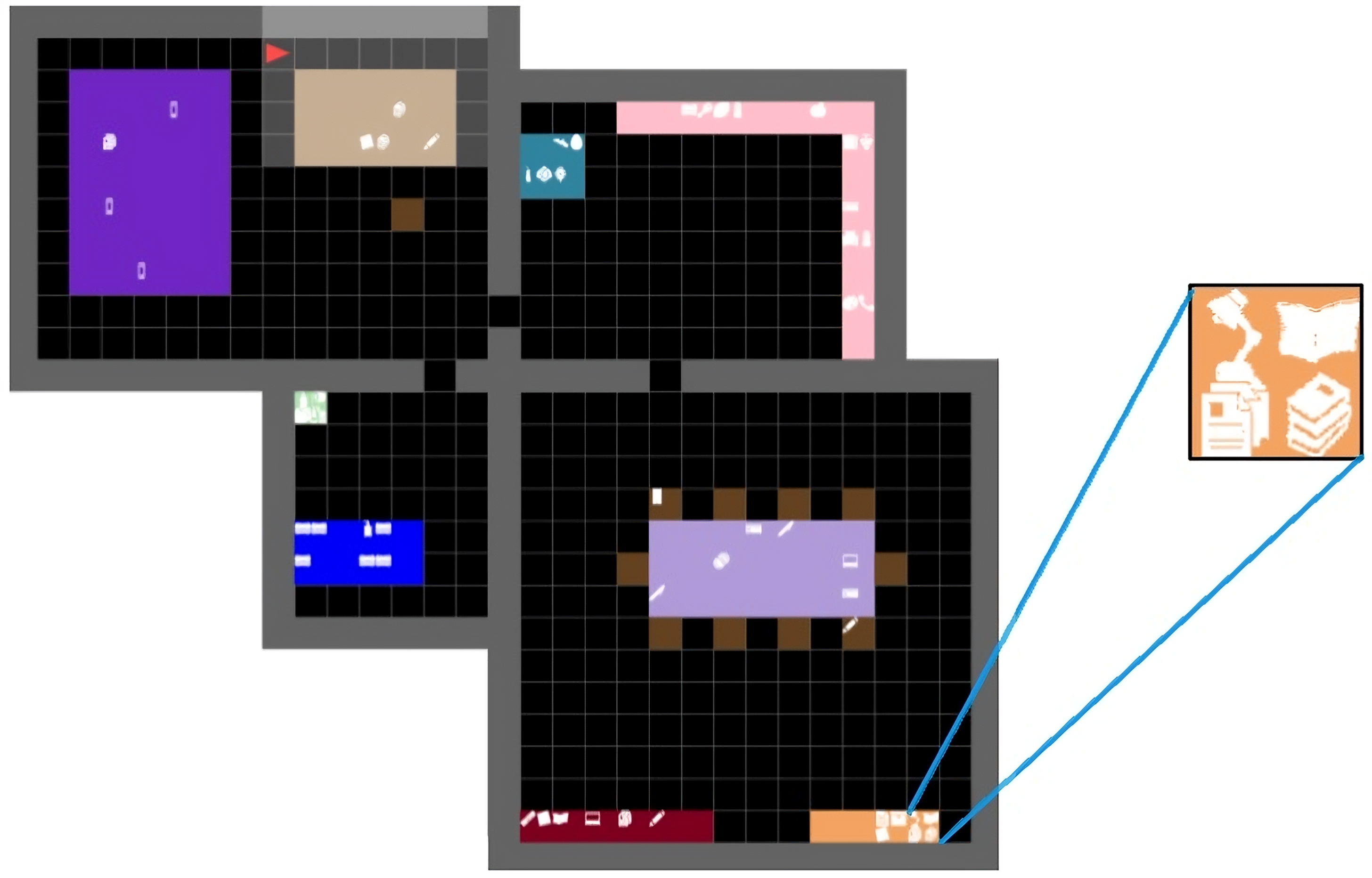}
    \caption{iGridson, our gridworld implementation of the iGibson 2.0~\cite{li2021igibson} simulator. An example household scene with four rooms is shown here with various furniture items and objects.}
    \vspace{-0.5cm}
    \label{fig:igridson}
\end{figure}

% \begin{figure*}
%     \centering
%     \includegraphics[width=1\linewidth]{figs/model.png}
%     \caption{Node Edge Selector (NES) model architecture. Please see Appendix TODO for details on node and edge features.}
%     \label{fig:model}
% \end{figure*}

\textbf{iGridson.} 
We implement a gridworld version of iGibson 2.0~\cite{li2021igibson}, named iGridson, to ground the agent in an environment with support for 3D-object relations. In particular, given an initial scene graph, iGridson instantiates an environment such that every object in the environment corresponds to a node in the scene graph, and the objects are sampled to satisfy all edge relationships in the scene graph. The environment evolves according to the object dynamics. More details are included in the Appendix (Sec. \ref{sa:igridson}).
%In between each timestep, the objects in the scene remain stationary, and the agent navigates in the environment and performs the object search task. The furniture and room positions are known to the agent, so it performs path planning (A* algorithm) to navigate to the selected furniture node. 

\subsection{Scene Graph Memory (SGM)}

As described in Sec.~\ref{sec:problem}, the goal of the agent is to predict the probability of query edges in each scene graph $SG_{t}^{i}$. 
%This corresponds to being able to construct a model of where objects are in the environment. 
Since each environment has distinct probability distributions $P^{i}$ and $T^{i}$, the agent needs to make use of the observations from the environment $O_0^i...O_t^i$ to estimate these environment-specific distributions. We propose the Scene Graph Memory (SGM) data structure to enable the agents to do so.

To simplify notation, the following applies to an agent operating in a specific environment $Env^i$. At timestep $t$, an instance of a Scene Graph Memory $SGM_t = (V^{SGM}_t, E^{SGM}_t)$ is composed of a set of nodes and edges of the same type as in the environment scene graphs. The SGM nodes $V^{SGM}_t = (\bigcup_{n=0}^{t} V^O_n)\bigcup V^Q_t$ are made up of all the observed nodes, as well as the set of query nodes -- the latter may not or may not be a subset of the former, since the query may be related to objects that agent has not seen before. The SGM edges $E^{SGM}_t = (\bigcup_{n=0}^{t} E^O_n) \bigcup E^H_t$ are made up of all the observed edges up until timestep $t$, and any new hypothetical edges. Hypothetical edges are edges the agent can predict for the query nodes without actually having observed them according to some function $E^H_t=f_h(V^Q_t)$; these are needed when dealing with queries of nodes that have not been observed yet or have few observed edges. $f_h$ could be implemented in various ways; we implement it by adding an edge to the SGM for every edge in $P^{prior}$ with a probability above a certain threshold. 
%Hypothetical edges are only kept for a single time step, as they are only needed to inform the possible locations of the query node of the current timestep.

Each node and edge in the SGM is associated with features reflecting the semantic properties of the object or relationship it represents as well its observed dynamics. Semantic properties are captured by word embedding associated with the node's rooms, furniture, or object label (eg ``mug''). We use the 96-dimensional Tok2Vec vectors optimized on \texttt{en\_core\_web\_sm} from the spaCy python package \cite{spacy2}. These semantic features are useful for recognizing nodes that are likely to share similar edges. The observed ``temporal'' features include the time since a node or edge has been observed, the number of times it has been observed, its state change frequency, and more. These observed features are necessary for the model to learn the environment-specific dynamics. More details are included in the Appendix (Sec. \ref{sa:sgm}).

A useful property of using the SGM representation is that it collapses the sequence of observations $O_0,...,O_t$ into a single graph. Thus, unlike prior works that relied on hard-to-optimize recurrent models, we only need our model to reason about $SGM_t$.

\subsection{Node Edge Predictor (NEP)}
\begin{figure}
    \centering
    \includegraphics[width=1\linewidth]{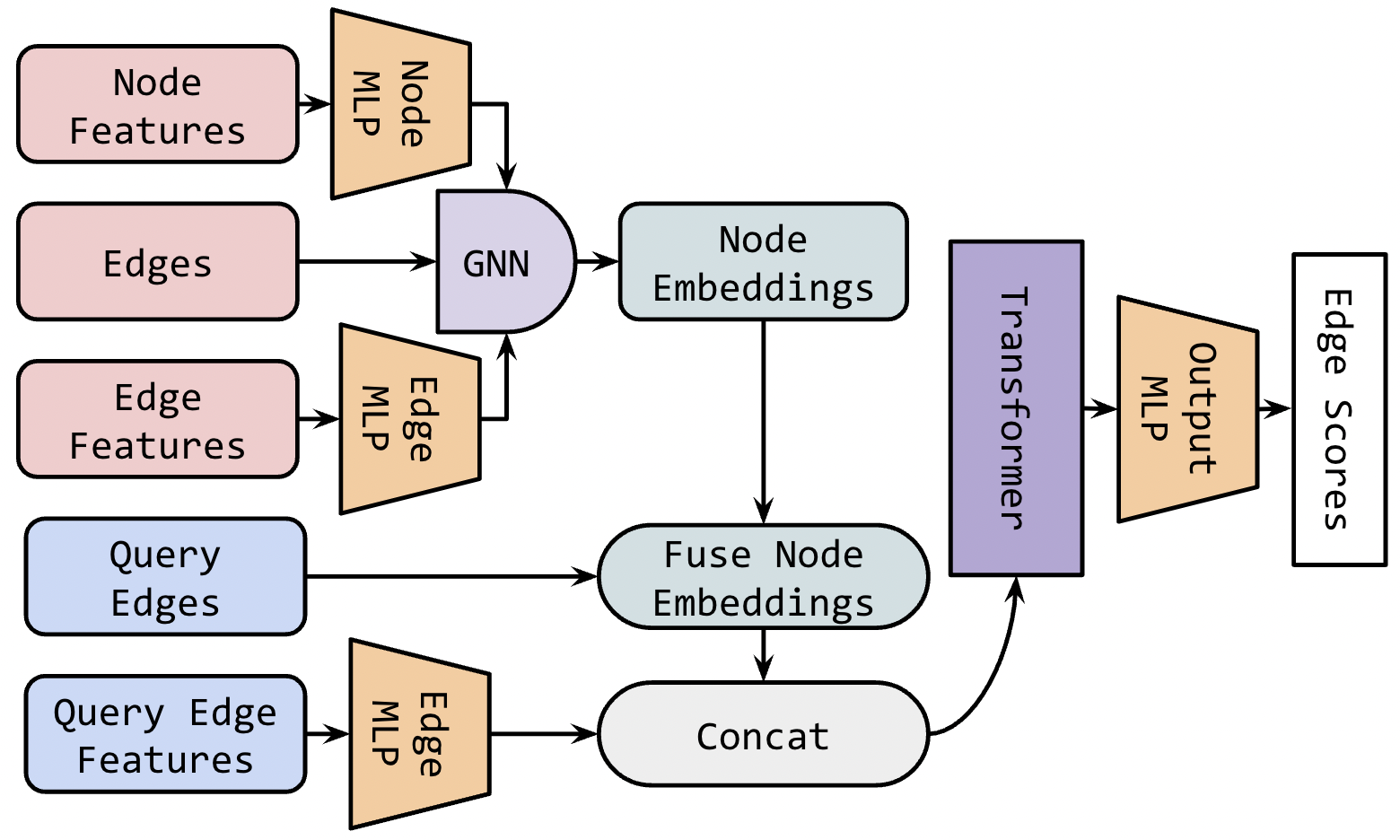}
    \caption{Node Edge Predictor (NEP) model architecture (GCN and HEAT variants). See the Appendix (Sec. \ref{sa:sgm}) for details on the node and edge features.}
    \label{fig:model}
\end{figure}
%Edges in the graph $SGM^i_t$ will in general be unknown, so we make the assumption that during data collection the agent can confirm or deny its predictions by further exploration. 
%The total number of SGMs in the dataset is $M*T$, where $M$ is the number of environments data collection is done in and $T$ is the number of timesteps data collection is done for in each environment.

\textbf{Model architecture.} We are concerned with predicting the probabilities of query edges, with a focus on the case when these edges all connect to a single node. This motivates our design for the Node Edge Predictor (NEP) model (Fig.~\ref{fig:model}).
%The need to compare the relative likelihood of query edges motivates several design decisions which we will now describe. 

NEP consists of four modules: node embedding, edge embedding, feature fusion, and edge classification. The node and edge embedding layers consist of a three-layer MLP with ReLU activations. The node embedding layers also optionally include graph convolutional layers after the MLP. %Graph neural net computations are done with message passing which allow varying graph sizes with different number of objects. 
% We choose to embed nodes before performing graph convolution because each node and edge so as to first process the raw input features and only then do message passing to propagate the embedded features.
In other words, we first embed the raw input features for nodes and edges and then optionally perform message passing to propagate the embedded features across the graph. 

We evaluate three variants of the model with different node embedding modules: \textbf{NEP-MLP} includes no graph convolution layers, \textbf{NEP-GCN} performs a two-layer graph convolution operation on top of the MLP embedding, and \textbf{NEP-HEAT} \cite{mo2021heat} replaces the graph convolution layers with heterogeneous edge-enhanced graph attentional (HEAT) operators, which condition the attention computed for message passing on both the node and edge features. \textbf{NEP-HEAT} is chosen as our primary variant because we expect that edge-conditioned attention will allow for more effective feature propagation. HEAT was originally designed for trajectory prediction of traffic, so this is the first time it is being applied to link prediction.
%  We also expect that having graph convolution on top of the MLP helps performance, as it allows for nodes with similar semantic types to ``share'' information as to where they have been.
% In other words, if the agent has seen oranges before, having GNN layers would enable it to make better predictions about apples.

After creating embeddings for all nodes and edges, we fuse the features of each pair of nodes associated with a query edge by averaging the nodes' embeddings and then concatenating that with the edge's embedding. Since different query nodes have different numbers of edges, we pad the input tensors for each batch before fusion. Lastly, the batch of fused features is passed to a 2-layer transformer encoder~\cite{vaswani2017attention}. The self-attention layers in the transformer enable the model to evaluate all the query edges jointly, which is important for cases in which a node's edges are mutually exclusive. NEP's use of a GNN followed by a transformer is most similar to GraphFormer from~\citet{yang2021graphformers} Heterformer from~\citet{jin2022heterformer}, but differs in that NEP uses the transfomer only on the embedded query edges, as opposed to the costlier use of transformers in alteration with GNN layers. Lastly, the transformer's output are passed through a 3-layer MLP with a Sigmoid activation in order to yield the output logits across all edge candidates. 

\textbf{Model training.} For each training iteration, we randomly sample a batch of SGMs from the training dataset, and batch all the query edges from each SGM.  The loss is computed by calculating the Binary Cross-Entropy between the logits and the ground-truth labels. We optimize for binary classification per edge as opposed to N-way classification because a given node may have multiple edges that are true. Because the number of true edges is far lower than false edges, we multiply the losses corresponding to false edges by the ratio of true edges to false edges before backpropagation. 

% \textbf{Model deployment} To leverage the model after training to adapt to new environments, an agent needs to continuously construct an SGM as it explores. When the agent receives a query node, it passes the SGM along with the associated query edges to the model, and then receives the posterior probability for each query edge. These can be used with a planning-based approach (i.e. always choose the furniture node of the most likely edge to explore next) or as input to another neural network for end-to-end online optimization. 

%% file: 5-exp.tex
\section{Experimental Setup}
\label{s:es}

We design our experiments to test whether the proposed \model models can outperform alternative approaches across three downstream tasks that involve link prediction. 

\subsection{Training Data Collection} 
In order to train the model, we first collect data by having an agent gather observations in a variety of training environments. The agent's goal is to complete the same object search task that the SGM model is meant to help with, and can follow any policy during data collection, such as the heuristic baselines explained below. As the agent tries the complete the task, it gathers observations and constructs SGMs along the way. Besides the features described above, each query edge in the $SGM^i_t$ is also annotated with a label -- true or false -- which corresponds to whether this edge is actually present in the ground-truth scene graph. Each node is also labeled as being a query node or not. Once a variety of SGMs with labels have been collected, they can be aggregated into an offline dataset for model training. 

% Our goal is to answer the following questions through these experiments:
% \begin{enumerate}
%     \item How well does our method perform as compared to different heuristics?
%     \item How do features of our model and methodology, such as the model's access to priors and textual labels about the object, impact the model's performance?
%     \item Lastly, can using different networks on top of a probabilistic scene graph improve performance?
% \end{enumerate}

\subsection{Tasks and Metrics}
We define three tasks with associated metrics as the basis of our experiments.

\textbf{Predict Object Location:} At every timestep, the agent must predict the location (furniture node) of an object with a particular description.  The agent is then able to observe the node it has predicted and its associated object nodes, regardless of whether its output is correct. The metric for the task is accuracy -- whether the agent correctly predicted a furniture node that is connected to an object with the correct description. %We consider a prediction correct if the prediction has edges to at least one node with a description that matches the query object.
This task is designed to evaluate how well an embodied agent may work in practice in an unknown environment; predicting the correct location of a given query object is essential for downstream tasks like household chores.

%Once the agent has received the observation, some objects are moved and the next step begins.

\textbf{Predict Relative Location Likelihood:}
At every timestep, the agent is queried for multiple objects and is required to predict the likelihood of each location that each object can be at. The metric for the task is the Normalized Discounted Cumulative Gain (NDCG), a popular method for measuring the quality of a set of search results\cite{jarvelin2002cumulated}. We choose this metric as we primarily care about the agent correctly predicting the ranking of the location options rather than the exact values in $T$. The agent is then able to observe the mostly likely nodes for each query node, as in the previous task. Compared to the previous task, this task and metric provide a more holistic evaluation of how well the agent models the entire environment at every timestep.

\textbf{Find Object:}
The prior two tasks involve a single prediction at each time step. This task has the same objective as predict location, but the agent is now allowed multiple sequential choices. After each location choice, the agent can observe the location and update its internal state before making the prediction for the next action. The environment is static during this, so exhaustive search will always succeed. The metric for this task is the number of actions it takes to pick a correct node.
%This task is designed to most closely mirror something an agent would need to do in a real household environment.

\subsection{Baselines} We compare our models with six heuristic-based models as well as the most highly relevant prior work. 

These are the heuristic baselines: 
\textbf{Random:} Randomly chooses an edge among all options.
\textbf{Frequentist:} Records the number of times edges have been observed to be true and false, and chooses the option with the highest ratio of true observations to total observations. 
\textbf{Priors:} Chooses the most likely option according to $P^{prior}$.
\textbf{Myopic:} Always chooses the last location each object was observed at, or at random if the object has not been observed.
\textbf{Bayesian:} Treats each edge in the SGM as having a distinct beta-binomial probability distribution. We create the distributions with a beta prior based on $P^{prior}$ and compute the posterior distribution by treating observations as a sequence of Bernoulli trials. More details in the Appendix (Sec. \ref{sa:task}).
\textbf{Oracle:} Uses ground truth knowledge about the dynamics of the scene as well as memory of past observed object locations to make the best choice possible.

Lastly, we also compare to the HMS model from \citeauthor{kurenkov2021semantic} This model was 
also designed for link prediction in the context of scene graphs, but in the context of static scenes. Thus, it is a good point of comparison for our revised problem definition. Compared to our model, it does not use the SGM representation and uses a standard GCN neural net rather than the NEP.
%\textbf{Oracle Upper Bound:} This model receives the location of all objects at the beginning of every scene and also has access to the likelihood with which objects can move in the scene. 

% \begin{table*}
% \centering
% \caption{\label{tab:predict_location_ablations} Model ablations.}
% \vspace{0.1cm}
% \resizebox{\textwidth}{!}{
%     \begin{tabular}{llllll}
%         \toprule
%         Model & Full Model & -Prior probabilities & -Temporal features & -Word vectors & -All \\
%         \midrule
%         PSG (MLP) & 0.241 & 0.206 & 0.254 & 0.250 & N/A \\
%         PSG (GCN) & 0.204 & 0.210 & 0.223 & 0.226 & N/A \\
%         PSG (HEAT) & 0.261 & 0.221 & 0.263 & 0.268 & N/A \\
%         \midrule
%     \end{tabular}
% }
% \end{table*}

%% file: 6-results.tex
\section{Experiment Results}
\label{s:er}
We evaluate our NEP variants against all baselines on the test set (unseen environments) for all three tasks. We train each NEP model with a dataset of 10,000 SGMs collected over 100 different environments, with 100 steps being taken per environment. The data collection is done by having the Bayesian baseline do the relevant task while creating and storing SGM graphs along the way.

%\begin{table}[h!]
%\centering
%\caption{\label{tab:train_test} Train and test accuracy for the NEP models. TODO cut? }
%\vspace{-0.1cm}
%\resizebox{\linewidth}{!}{
%    \begin{tabular}{l|ccc}
%        \toprule
%         & NEP-MLP & NEP-GCN & NEP-HEAT \\
%        \midrule
%         Train & 0.242 &  0.247  & 0.247  \\
%         \ Test & 0.244 & 0.237 & 0.243 \\
%        \midrule
%    \end{tabular}
%    }
%\end{table}

%L2 error is computed between the logits and the ground truth error and averaged across the entire train and test set. In general, all models suffer from a degree of overfitting, which is more so the case for the GNN models. TODO detailed training results are included in the appendix.
%As shown in table \ref{tab:predict_location_results}, the GNN models having greater L2 test error does not appear to correspond to worse performance on downstream tasks. Nevertheless, it is possible that models with GNNs may benefit from additional regularization.

\begin{table*}[t!]
\centering
\caption{\label{tab:results} The mean accuracy and standard deviation (averaged across 100 runs/environments) are shown for three tasks and different environment conditions. Lower is better for Predict Object Location and Find Object. Dynamic nodes refer to whether nodes are added and removed throughout scene evolution, which makes the problem significantly more challenging. \\ }
\vspace{0.1cm}
\resizebox{0.95\textwidth}{!}{
    \begin{tabular}{c|cc|cc|cc}
        \toprule
        Task & 
        \multicolumn{2}{c|}{Predict Object Location $\uparrow$} &
        \multicolumn{2}{c|}{Predict Relative Location Likelihood $\uparrow$} &
        \multicolumn{2}{c}{Find Object $\downarrow$}
        \\
        \cmidrule(lr){1-1}\cmidrule(lr){2-3}\cmidrule(lr){4-5}\cmidrule(lr){6-7}
        Dynamic Nodes? & \multicolumn{1}{c}{Y} & \multicolumn{1}{c|}{N} & \multicolumn{1}{c}{Y} & \multicolumn{1}{c|}{N} &  \multicolumn{1}{c}{Y} & \multicolumn{1}{c}{N}\\ 
        \midrule
    Random     & 0.046 $\pm$ 0.005     &     0.044 $\pm$ 0.005     &      0.381 $\pm$ 0.001     &    0.384 $\pm$ 0.001 & 8.769 $\pm$ 1.092  & 8.750 $\pm$ 1.100  \\
    
    Priors     & 0.159 $\pm$ 0.019    &    0.180 $\pm$ 0.025     &     0.536 $\pm$ 0.001    &    0.546 $\pm$ 0.002    &    5.911 $\pm$ 1.433 & 5.869 $\pm$ 1.438    \\

    Frequentist     & 0.271 $\pm$ 0.029     &    0.315 $\pm$ 0.038    &     0.668 $\pm$ 0.001 &    0.728 $\pm$  0.002  &  4.550 $\pm$ 1.571 & 4.218 $\pm$ 1.624   \\
    
    Myopic     & 0.282 $\pm$ 0.030    &    0.339 $\pm$ 0.043     &     0.029 $\pm$ 0.000    &     0.017 $\pm$  0.000 & 6.333 $\pm$ 2.202 &  6.048 $\pm$ 2.186   \\
    
    HMS   &     0.194 $\pm$ 0.023     &     0.210 $\pm$ 0.033     &      0.529 $\pm$ 0.001     &     0.582 $\pm$ 0.002 & 5.357 $\pm$ 1.379 &  5.254 $\pm$ 1.458 \\
    
    Bayesian     & 0.286 $\pm$ 0.028    &    0.308 $\pm$ 0.034     &     0.699 $\pm$ 0.002    &  0.725 $\pm$ 0.0 02 &  3.469 $\pm$ 0.761 & 3.610 $\pm$ 0.970 \\
    
    \textbf{NEP-MLP}    & 0.302 $\pm$ 0.032     &     0.336 $\pm$ 0.038     &   0.661 $\pm$ 0.002 &  0.736$\pm$ 0.002  & 3.650 $\pm$ 1.035 & 3.445 $\pm$ 0.948    \\

    \textbf{NEP-GCN}    &  0.324 $\pm$ 0.034     &     0.359 $\pm$ 0.041     &    \textbf{0.728 $\pm$ 0.001} &     \textbf{0.792 $\pm$ 0.002}  & 3.629 $\pm$ 1.042 & 3.191 $\pm$ 0.866    \\
    
    %NEP-HAN   & TBD $\pm$ TBD     &     TBD $\pm$ TBD     &     0.150 $\pm$ 0.021    &     0.229 $\pm$ 0.027 & TBD  $\pm$ TBD & TBD  $\pm$ TBD \\
    
    %NEP-HGT   & TBD $\pm$ TBD     &    TBD $\pm$ TBD     &    0.211 $\pm$ 0.025     &      0.265  $\pm$ 0.029  & TBD  $\pm$ TBD & TBD  $\pm$ TBD    \\
    
    \textbf{NEP-HEAT}   &  \textbf{0.351 $\pm$ 0.033}    &     \textbf{0.391 $\pm$ 0.041}     &     0.724 $\pm$ 0.002 &     0.783 $\pm$ 0.002 & \textbf{3.570 $\pm$ 1.032} & \textbf{3.186 $\pm$ 0.910} \\
    
    \midrule
    Oracle     & 0.454 $\pm$ 0.039     &   0.475 $\pm$ 0.063     &     0.932 $\pm$ 0.002     &    0.939 $\pm$  0.002 & 3.001 $\pm$ 0.772 & 2.921 $\pm$ 0.764  \\
        \midrule
    \end{tabular}
}
    \vspace{-0.5cm}
\end{table*}

\subsection{Task Performance}

\begin{figure}[h!]
    \centering
     \includegraphics[width=0.9\linewidth]{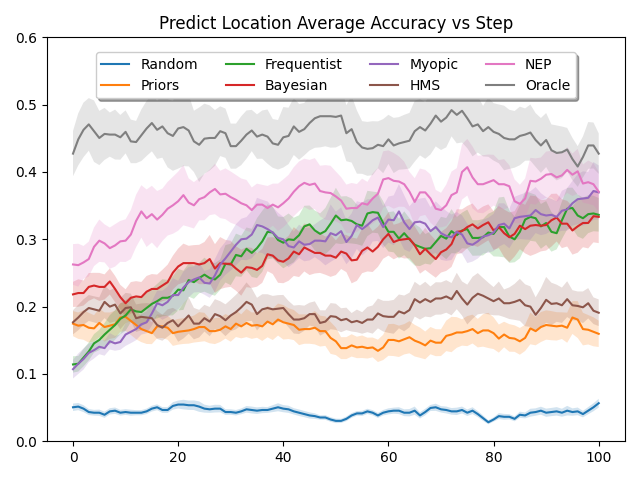}
     \vspace{-0.05cm}
     \caption{The average accuracy and variance for the Predict Object Location task, averaged across 100 different environments. The x-axis is the environment steps. The y-axis is smoothed with averaging over 10 steps. NEP results are with the HEAT variant.}
     \label{fig:main_result}
     \vspace{-0.1cm}
\end{figure}

The main results can be seen in Fig.~\ref{fig:main_result} and Table~\ref{tab:results}. We have the following observations. 

First, as seen in Fig.~\ref{fig:main_result}, the use of SGM enables the NEP to improve over time, unlike models that completely rely on the prior information and do not make any use of observations (Priors and HMS). The model also outperforms the baselines that do make use of observations (Frequentist, Myopic, and Bayesian), which we hypothesize is due to the NEP's understanding of object semantics, use of GNN-based feature propagation, and the self-attention mechanism that allows for edge comparisons.

Second, the priors embedded in the SGM as well as learned semantic patterns make the NEP able perform better than all heuristics from the outset. This gives NEP a head start compared to the models that purely rely on observations made during test time (Myopic and Frequentist). These baselines do approach the performance of NEP over time, since eventually it's possible to model the dynamics perfectly. However, NEP still has the advantage of faster and better adaptation to new environments. 

Third, NEP also significantly outperforms the Bayesian method, which is given access to both the priors and observations over time. We believe this is because the Bayesian approach reasons on a per-edge basis, whereas NEP is able to propagate information about edge features with the use of a GNN and fuse information about query edges with the use of self-attention. Thus, it is possible to share information about the dynamics of nodes with similar semantics. 

Lastly, the detailed results in Table~\ref{tab:results} demonstrate several things. As could be expected, making the environment dynamic both in terms of object location and object presence makes the tasks harder, compared to always dealing with the same set of objects. The HEAT NEP variant generally performs best, but the simpler GCN variant performs slightly better on the predict relative likelihood task, possibly because producing a ranking of edges is easier than specifically predicting the most likely edge. Lastly, the first two tasks do correspond to better downstream performance on finding objects through a sequence of actions, as is our ultimate goal.

The results support our initial hypothesis that combining the proposed representation (SGM) with the NEP model is suitable for the temporal link prediction tasks in our setting, as this combination allows generalization to unseen environments as well as online adaptation.

\subsection{Ablations}
We further perform several ablation studies to test the relative importance of the components in our model (Fig.~\ref{fig:model}). The results are shown in Table~\ref{tab:predict_location_ablations}. It is evident that all of our model designs contribute significantly to the model performance. The prior probability information is most critical, not only because it is included in node edge, but crucially also because it is used for selection of hypothetical edges; with no priors, hypothetical edges are sampled at random and so may not include the true object location. Even though the priors do not match the test environments' true dynamics, they are highly beneficial for improving performance. 

The use of the transformer layers, a key aspect of the NEP model, turns out to be the second most important design choice. This validates our hypothesis that it is useful to perform self-attention over query edges prior to predicting their likelihoods. Interestingly, temporal features are the most important for the HEAT variant, suggesting it makes the most use of them for adapting to novel environments. Lastly, semantic features in the SGM turn out to be the least important, perhaps because temporal features alone are sufficient to infer semantics.

\begin{table}
\centering
\caption{\label{tab:predict_location_ablations} Ablation study results. The mean accuracy and variance (averaged across 100 runs/environment) are reported for for the Predict Object Location task with dynamic nodes. Higher is better. }
\resizebox{\linewidth}{!}{
    \begin{tabular}{l|ccc}
        \toprule
         & NEP-MLP & NEP-GCN & NEP-HEAT \\
        \midrule
         Full Model & 0.302 $\pm$ 0.032 & 0.324 $\pm$ 0.034 & 0.351 $\pm$ 0.033 \\
        \ (-) Prior probability & 0.260 $\pm$ 0.028 & 0.233 $\pm$ 0.024 & 0.267 $\pm$ 0.029\\
         \ (-) Transformer & 0.263 $\pm$ 0.028 & 0.310 $\pm$ 0.034 & 0.314 $\pm$ 0.032 \\  
         \ (-) Temporal features & 0.288 $\pm$ 0.032 & 0.320 $\pm$ 0.033& 0.322 $\pm$ 0.031 \\
         \ (-) Semantic features & 0.263 $\pm$ 0.027 & 0.322 $\pm$ 0.030 & 0.328 $\pm$ 0.030 \\
        \midrule
        \end{tabular}
    }
    \vspace{-0.2cm}
\end{table}

\subsection{Downstream Task Performance in iGridson}
We also evaluate the agent's performance on the Find Object task in the iGridson environment shown in Fig~\ref{fig:igridson}. The quantitative results are shown in Table~\ref{tab:igridson_results}. This environment contains only 21 furniture locations, so the agent is successful if it predicts the correct location by searching up to roughly half the options. Unlike the Find Object in Table~\ref{tab:results}, the iGridson representation includes a spatial layout rather than just a symbolic one. Therefore, it is possible to compute the path length the agent traverses across all its actions. The NEP method is always able to find the object, and does so within fewer actions and shorter paths. The variances for these metrics are surprisingly high, potentially because some objects are much harder to find than others. We hope to explore performance in iGridson more in the future.

\begin{table}
\centering
\caption{\label{tab:igridson_results} Mean success rate, number of actions, and path length ($\pm$ standard deviation) to reach the query objects in the iGridson environment averaged over 10k runs. The agent is given at most 10 actions to reach the target object. 
}
\resizebox{\linewidth}{!}{
    \begin{tabular}{c|ccc}
        \toprule
         & Success Rate & \# of Actions & Path Length \\
        \midrule
           Random   & 0.57 & 9.36 $\pm$ 4.09  & 141.74 $\pm$ 63.86\\
         \ Myopic   & 0.57 & 9.31 $\pm$ 4.14  & 140.92 $\pm$ 64.40 \\
         \ Bayesian & 0.84 & 5.06 $\pm$ 4.50  & 70.90 $\pm$ 63.06\\
         \ \textbf{NEP-HEAT} & \textbf{1.00} & \textbf{1.56} $\pm$ \textbf{0.86}  & \textbf{30.05 $\pm$ 24.67}\\
        \midrule
    \end{tabular}
    }
\end{table}

%% file: 7-discuss.tex
\section{Conclusion}
We proposed a new problem setup, temporal link prediction for dynamic and partially observable graphs, as well as a first solution to this problem in the context of object search in embodied AI. Our work includes a new benchmark, the Dynamic House Simulator, where it is possible to evaluate model performance on dynamic link prediction in the embodied AI context. Our solution to predict object locations and environment dynamics efficiently includes a new representation, Scene Graph Memory (SGM), and a novel net learned architecture, the Node Edge Predictor (NEP). Our SGM-based models significantly outperform all the alternative approaches due to their ability to (1) learn scene statistics (commonsense knowledge about object relations) during training, and (2) adapt online by leveraging noisy, partial observations. These features help embodied AI agents perform object searches in unseen, dynamic, and partially observable environments, as well as link prediction models with partially observable graphs in general. 

%% file: 8-appendix.tex
\appendix
\section{Dynamic House Simulator}
\label{sa:dhs}

\begin{figure*}[h!]
    \centering
    \includegraphics[width=\linewidth]{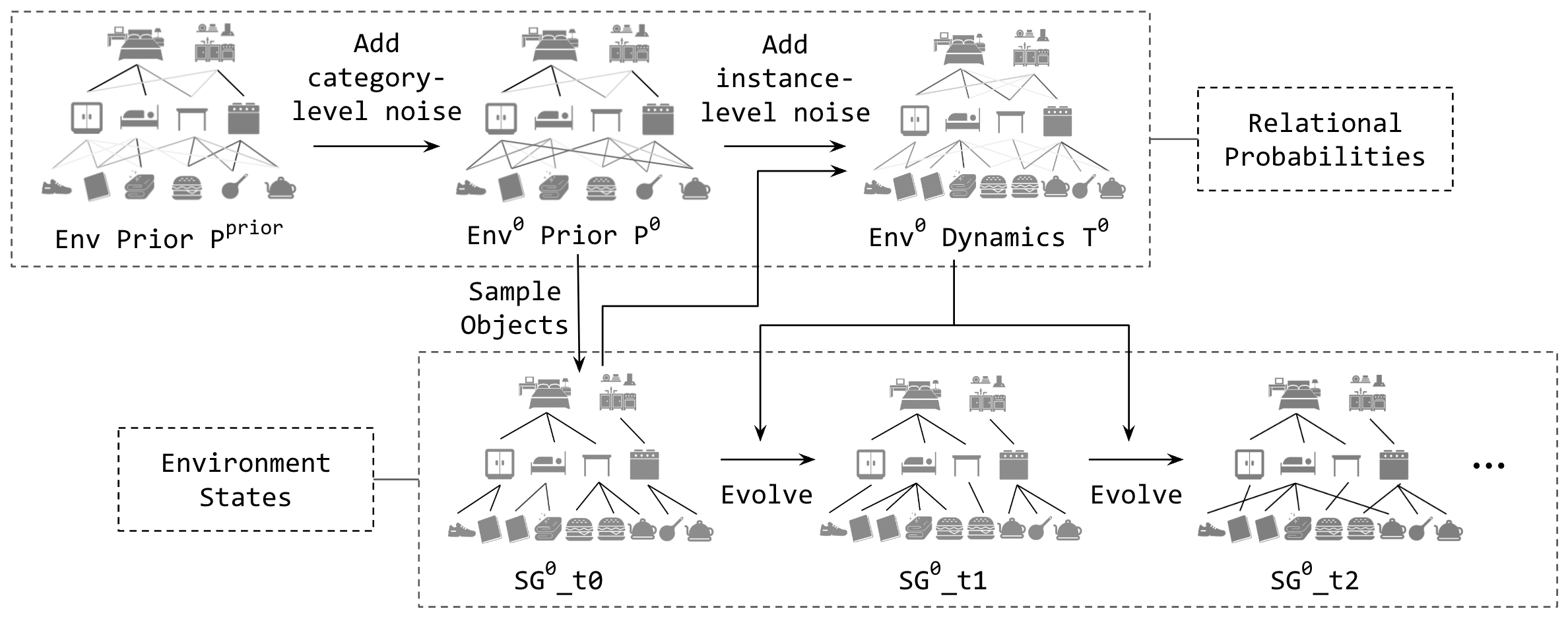}
    \caption{Illustration of Dynamic House Simulator scene Sampling and Evolving process. As outlined in Sec.~\ref{sec:problem} and Algo.~\ref{alg:dhs}, for each environment $Env^{i}$, we first inject object class-level noise $N_{class}^{i}$ to the global prior $P^{prior}$ to generate environment-specific relation probabilities $P^i$. The initial scene graph $SG_{0}^{i}$ will be sampled based on $P^i$ in a procedural manner. Then we inject object instance-level noise $N_{instance}^{i}$ into the environment-specific relation probabilities $P^i$ to generate the environment dynamics $T^i$, which will evolve the scene graph across time steps.}
    \label{fig:dhs}
\end{figure*}

A visual representation of Alg.~\ref{alg:dhs} is shown in Fig~\ref{fig:dhs}. 

\begin{figure*}[h!]
    \centering
    \includegraphics[width=0.98\linewidth]{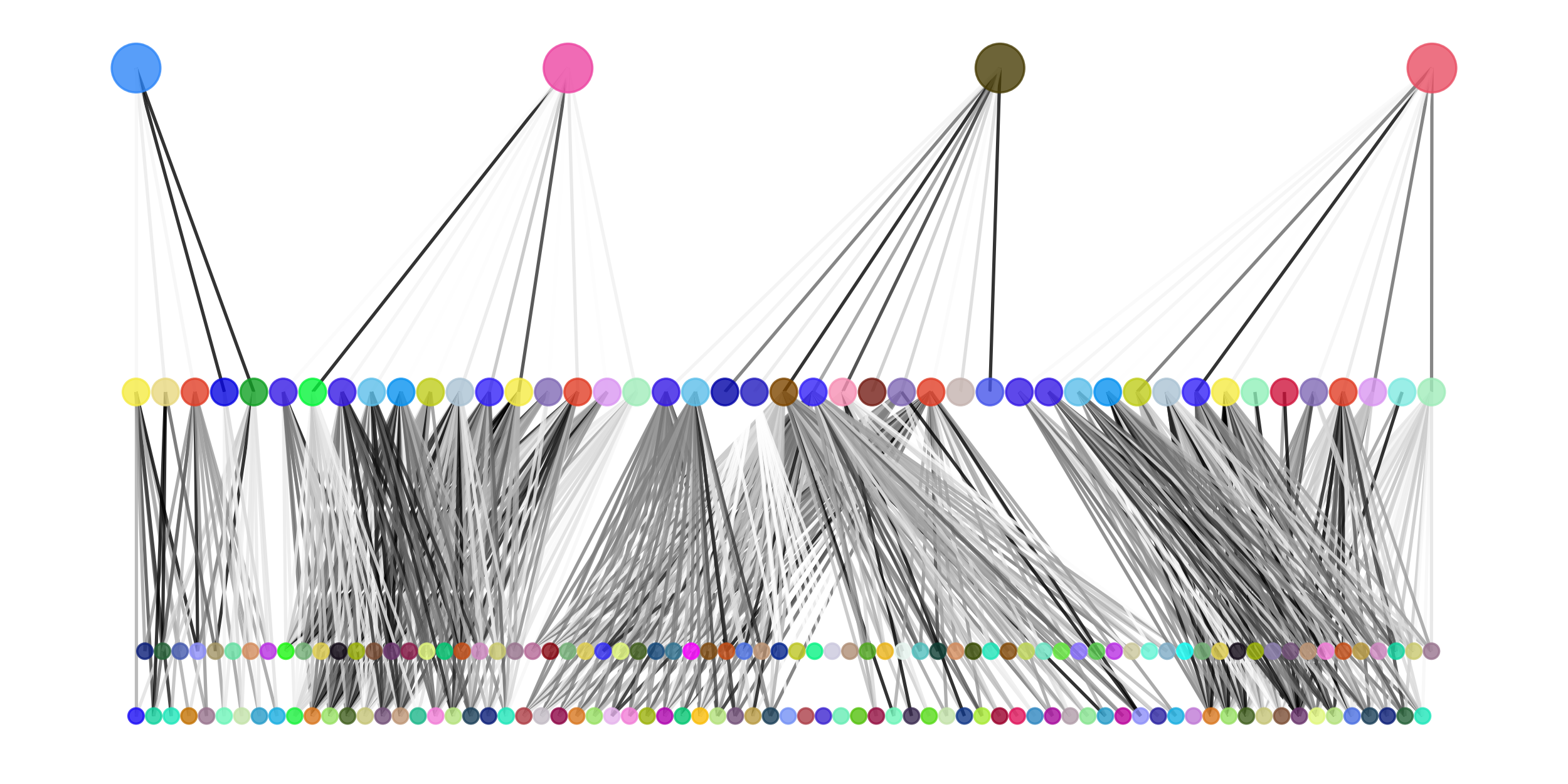}
    \caption{The household object placement probability priors. Nodes at the top correspond to rooms, nodes at the middle correspond to furniture, and nodes at the bottom correspond to objects. Node color a unique room, furniture, or object label. Edge color represents the likelihood of a piece of furniture being sampled for a room or the likelihood of an object being sampled for a piece of furniture, with darker edges having a higher likelihood. To improve visualization, a node with a given label is created per room that it is connected to.}
    \label{fig:priors_graphs_detailed}
\end{figure*}

\textbf{Object placement probabilities}
 For ProcThor-10k, we counted the occurrences of object relationships and normalized to get point probabilities. iGibson already includes the probabilities as metadata, so we use that directly. For relationships that are in both, we use the ProcThor-10k. We compute object placement probabilities for ProcThor-10k by counting the number of occurrences in each relationship type within the 10,000 published train environments and normalizing the counts of all room-furniture relationships and furniture-object relationships. Specifically, we create counts of furniture-object edges counts per room and divide by the sum of all object instance in each room. iGibson 2.0 already encodes such probabilities in its metadata, so we only need to average the F-O probabilities for the coarse graph. We filter out any furniture nodes with fewer than 3 outgoing edges to focus computation on simulating busier aspects of the environment. The resulting $P_{prior}$ graphs is visualized in Fig. \ref{fig:priors_graphs_detailed}.

\begin{figure*}[h!]
    \centering
    \includegraphics[width=0.98\linewidth]{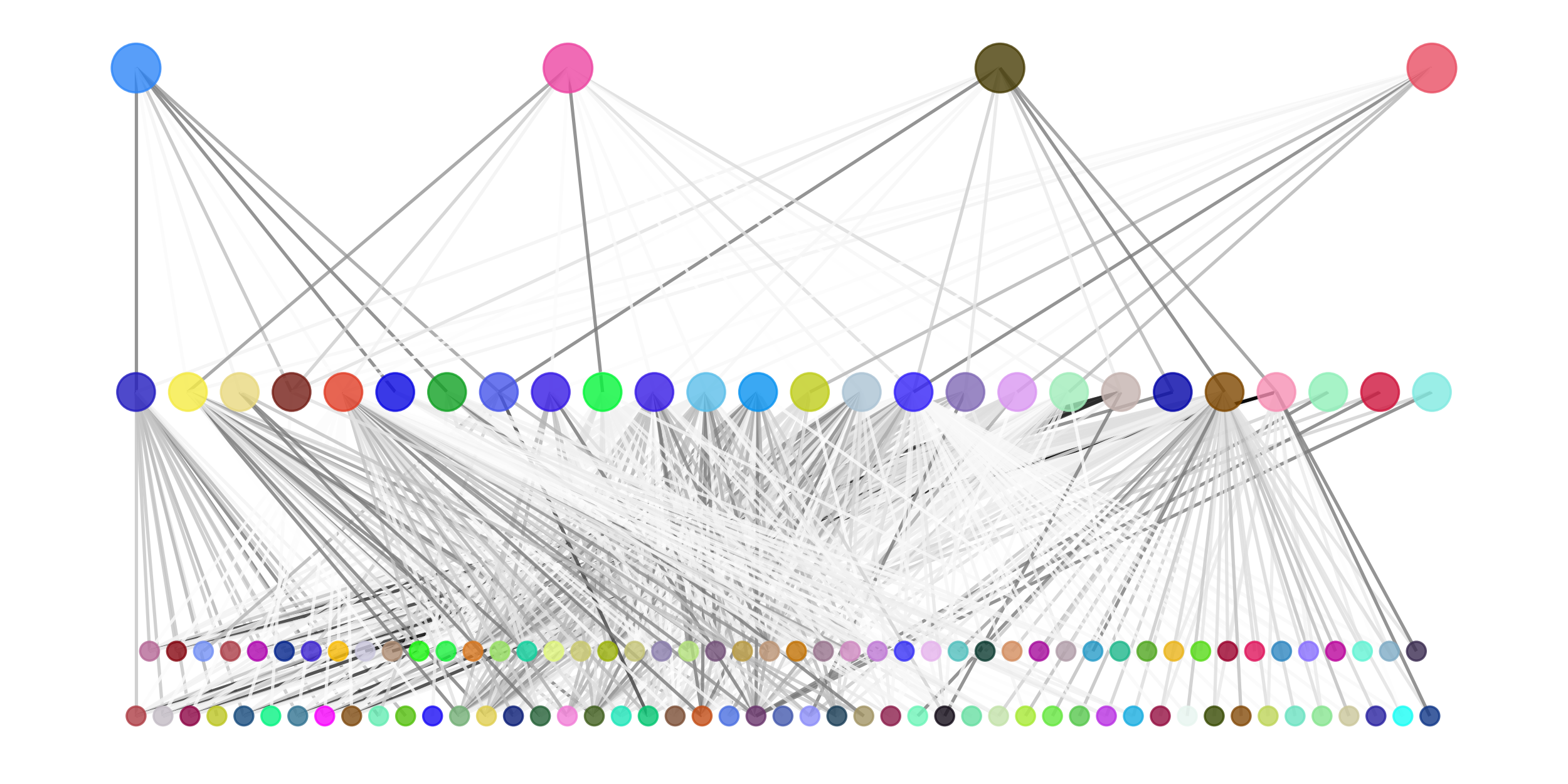}
    \caption{The ``coarse'' household object placement probability priors. The layout and use of color is the same as in Fig. \ref{fig:priors_graphs_detailed}. In the this version of the priors, the furniture-object relationships (i.e. edges at the bottom level) are independent from the room-furniture relationships (i.e. edges on the top level). Nodes with a given label are not replicated per room, meaning that each node represents a distinct room, furniture, or object type. Some nodes have primarily white edges because they have a large number of outgoing edges, which makes the probability of each edge lower compared to nodes with few outgoing edges. }
    \label{fig:priors_graphs_coarse}
\end{figure*}

 We also create a the ``coarse'' version of the priors graph in which furniture-object edges are counted irrespective of rooms and divided by the sum of all object instances.  The resulting $P_{prior}$ graphs is visualized in Fig. \ref{fig:priors_graphs_coarse}. We did not use these priors in our experiments, since they lead to object dynamics that are strictly simpler than the ``detailed'' priors and therefore the results were less informative.
 
\textbf{Sampling noise}
There are two types of class-level noise: sparsification and randomization. Sparsification means we zero out probabilities for some edges, e.g. a certain object will never be inside a certain piece of furniture. If an edge's probability is not zeroed out, we then randomly scale it up or down by a certain amount. After the application of noise, all edge probabilities for the object are normalized. The goal is to make each environment instance have its own specific locations for various objects. In our experiments we use a probability of 25\% for both zeroing out edges and as the maximum and minimum by which probabilities may be scaled.

\textbf{Environment Sampling}
An environment consists of 1 floor, 4 rooms (kitchen, living room, bedroom, bathroom), 8 furniture items per room, and 6 objects per piece of furniture. While our approach is compatible with different counts for each category, we chose this specific ratio (floor:room:furniture:item) to standardize our experiments. The furniture and object items are sampled from the candidate lists in Table~\ref{tab:sampleable_furniture} and Table~\ref{tab:sampleable_objs}. Adjectives are additional descriptors belonging to an adjective category such as size or color which are then sampled and attached to both furniture and objects by the following procedure. First, the number of adjective categories to attach to a given object or furniture items is sampled from a uniform distribution between 1 and the maximum number of categories. Then a specific adjective is uniformly sampled for each of these categories. This can result in multiple distinct objects sharing the same class existing in the scene, such as a "small green bag" or a "large red bag".
 
\textbf{Evolution of the environment}
The evolution of the environment is determined by the move frequency and add/remove probability outlined in Table~\ref{tab:sampleable_objs}. First objects are removed based on the remove probability, with a constraint that the object count cannot dip below 95\% of the original object count. In the second step, 5\% of all objects are moved. The distribution of which objects move is determined by the move frequency of the individual move probabilities based on the object class as outlined in Table~\ref{tab:sampleable_objs}. Finally, we iterate over every possible object and spawn objects according to the add probability with the constraint that the total number of objects in the scene cannot be higher than 105\% of the initial object count.

\section{SGM representation}
\label{sa:sgm}
 The SGM consists of a set of node ($V^{SGM}_t$) and edges ($E^{SGM}_t$). The node features are the following:
 
\begin{itemize}
    \setlength\itemsep{0em}
    \item The text embedding for the node is the 96-dimensional Tok2Vec vectors optimized on
\texttt{en\_core\_web\_sm}  from the spaCy python package \cite{spacy2}: 96 dimensional vector of floats in the range 0.0-1.0.
    \item The number of timesteps since the node was last observed: integer scalar in the range 0-100 (max timesteps).
    \item The number of times the node was observed: integer scalar in the range 0-100.
    \item The number of timesteps since the object was observed to have moved, which occurs during an evolution of the environment. This requires the agent to have observed the object to at least twice: integer scalar in the range 0-100.
    \item The observed frequency for which an object  has moved: scalar with value in the range 0.0-1.0.
    \item The node type ('house', 'floor', 'room', 'furniture', 'object'): 5 dimensional 1-hot vector.
\end{itemize}

The edge features are the following:

\begin{itemize}
    \setlength\itemsep{0em}
    \item The cosine similarity between the Tok2Vec text embedding of the edge's two nodes: scalar in the range 0.0-1.0.
    \item The number of timesteps since the edge was last observed: Integer scalar in the range 0-100.
    \item The number of timesteps since the observed last state change: scalar with the value 0.0 or 1.0.
    \item The number of times the edge was observed: integer scalar in the range 0-100.
    \item The number of times the edge state was observed to be true (present in the scene graph): integer scalar in the range 0-100.
    \item The frequency with which the edge observed to be  true (present in the scene graph): scalar in the range 0.0-1.0.
    \item The number of times the edge state was observed to have changed: integer scalar in the range 0-100.
    \item The last observed state of the edge: scalar with value 0.0 or 1.0.
    \item The prior probability that the edge exists derived from the ProcThor-10k and iGibson datasets: scalar with value in the range 0.0-1.0.
    \item The edge type ('in', 'contains', 'onTop', 'under'): 4 dimensional 1-hot vector.

\end{itemize}

All features are concatenated to form the feature vectors of nodes and edges. Prior to concatenation, temporal features (integers in the range 0-100) are first normalized to be roughly in the range 0.0-1.0 via multiplication by a manually derived scaling factor. For nodes this factor is $average\_scene\_graph\_num\_nodes (200) / (steps\_per\_scene (100)*10.0)$, and for edges this is $average\_scene\_graph\_num\_edges (500) / (steps\_per\_scene (100)*10.0)$. 

\section{Training hyperparameters}
\textbf{Implementation} All models are implemented and trained using PyTorch~\cite{pytorch} and PyTorch-Geometric~\cite{pyg}. Unless otherwise stated, we use the default parameters of these two packages.

\textbf{Model parameters} The node embedding network and edge embedding networks are two-layer feedforward neural networks with 64 units and ReLU activations following each layer. The HEAT and GCN graph neural networks have one graph convolution layer with 64 units. The transformer encoder is a standard self-attention model with 2 heads with a 64 unit feedforward network. 

\textbf{Training parameters} The model was trained for 25 epochs with a batch size of 100. The Adam optimizer was used with a learning rate of $1\times10^{-4}$.

\section{Task and Baseline Implementation Details}
\label{sa:task}
\textbf{Detection error} For all tasks, we simulate detection error by randomly skipping 25\% of the objects during observation.

\textbf{Choice of query object} For the predict object location task, the query object is sampled at random either from the set of objects that has moved since the last step or from all object nodes. In the former case, the node is sampled uniformly, and in the latter case, it is sampled with a weight proportional to its movement probability. This ensures the agent sometimes has to predict the location of an object that is guaranteed to have been moved since it was last observed, which is a capability we wish to be able to evaluate. 

\textbf{Bayesian policy}  The bayesian baseline is based on a Bernoulli model with a beta prior. The beta prior parameters are computed to match a desired variance. Given prior probability $\mu$ and desired variance $v$, the compute the following:

\begin{algorithmic}
\STATE $\alpha \gets \mu^2 \times \left(\frac{1 - \mu}{v} - \frac{1}{\mu}\right)$
\STATE $\beta \gets \alpha \times \left(\frac{1}{\mu} - 1\right)$
\end{algorithmic}

This variance is chosen to be relatively large (0.05) to reflect that the priors don't in general match the dynamics of environments. The posterior predictive for an edge being true is then computed as follows:

\begin{algorithmic}
\STATE $\alpha_n \gets \alpha + \sum_{n=1}^{N} x_n$
\STATE $\beta_n \gets N - \sum_{n=1}^{N} x_n + \beta$
\STATE $p \gets \frac{\alpha_n}{\alpha_n + \beta_n}$
\end{algorithmic}

\section{Additional Discussion}
\subsection{Combining our method with Reinforcement Learning}
Multiple approaches have been proposed for integrating scene graphs with RL agents for navigation tasks, and as scene graph memory is the same sort of data structure it can be used in the RL scenario in the same sorts of ways. Examples of such approaches include~\cite{ravichandran2022hierarchical,li2021ion,seymour2022graphmapper,pal2021learning}. As in these works, the SGM data structure can be used as input to the RL agent alongside its raw observations to aid the agent’s decision making.
Further, these works all utilize GNN layers to process the scene graphs as part of the policy network, and our NEP architecture could be used to fulfill this purpose. We performed some exploratory experiments in training an RL agent in the iGridson environment, and found convergence to be non-trivial. Therefore, we leave this problem to future work.

\subsection{The complexity of Dynamic House Simulator tasks}

While our simulator may seem simple, the environments the simulator generates are in many ways more complicated than any existing benchmarks for object search. Concretely, our generated environments are complex in terms of:
\begin{itemize}
\item Diversity of scenes – most benchmarks in this space only support a set number of pre-generated scenes, whereas ours can generate endless scene  variations through controllable sampling
\item Size of scenes – per Appendix A, our experiments are conducted in scenes with 4 rooms, 32 furniture items, and 192 objects. Prior benchmarks are typically limited to smaller spaces with significantly fewer objects.
\item Variety of furniture and object types - our simulator supports over 20 furniture types and over 100 object types, with most of them having 3 or more possible modifying adjectives. This is far larger than the furniture or object variety in other benchmarks.  
\item Dynamics evolution - no other benchmark supports continual evolution of the scene state.
\end{itemize}

While our experiments focus on demonstrating results for symbolic scene graphs and the semi-realistic 2D iGridson environment, the approach we take for the latter can be directly extended to object search in realistic 3D spaces, as discuss next.

\subsection{Applicability of our approach to realistic 3D embodied object search}
Our work abstracts away the challenges of perception and navigation that are part of embodied object search in order to focus on the problem of modeling dynamics with the use of memory. Complementary to our focus, many recent works have demonstrated impressive performance on embodied instances of this problem that do require robust perception and navigation, but do not require modeling of environment dynamics of long term memory. A possible future direction for our research is to demonstrate the usefulness of scene graph memory and the node edge predictor model in the context of realistic embodied instances of this problem that these works address.

One possible approach for doing this is a direct extension of our approach for implementing the iGridson agent discussed in section 6.3 and Appendix F. Analogous to our 2D iGridson agent implementation, an agent can continually maintain a scene graph memory via processing of embodied observations, and then use NEP predictions to decide on navigation goals, which can be achieved via robust point-goal navigation. The works cited above could likely form the basis for implementing the necessary perception and point-goal navigation. In particular, the recent paper “Long-term object search using incremental scene graph updating”~\cite{zhou2022long} demonstrated the feasibility of incremental scene graph updating from visual observations, which can be directly extended to incrementally building scene graph memory.

Our simulator can also form the basis for adding dynamic objects to the benchmarks used in prior works~\cite{trivedi2019dyrep,kipf2019contrastivemchangcompositional,yi2020clevrer, bear2020learning,zhou2022long}. As we demonstrate with the iGridson component of our simulator, it is possible to translate the symbolic scene graphs of the environment into embodied spaces. While we only demonstrate this for the simplified 2D setting, the same approach can be extended to more realistic 3D simulators by sampling object placements according to the scene graph. For instance, iGibson 2.0~\cite{li2021igibson} supports sampling object placements to satisfy object placement constraints. The primary difficulty with implementing this would be acquiring sufficient 3D assets for all the objects our simulator supports.

\section{iGridson Embodied Environment Details}
\label{sa:igridson}
The iGridson simulator is a 2D embodiment of a home environment laid out in a grid format, built using the minigrid library~\cite{minigrid}. It enables translating the scene graphs generating by the Dynamic House Simulator into a spatial arrangement, which is what embodied agents actually have to deal with. 

As shown in Fig.\ref{fig:igridson}, the environment has a fixed layout of the home with four rooms, each with a predetermined number and type of furniture objects. In total, the environment consists of 21 furniture objects spread across different rooms, all of which remains static throughout the evolution of the environment. A set of objects are then placed on top of these furniture objects, and are dynamically moved to a different furniture at each time step based on their priors and specific dynamics as described in section 4.1. Note that in our experiments, the overall set of objects itself is static. That is, no new objects are added to the environment, nor are any existing objects removed, they simply move from one furniture object to another.

Our experiments in this environment follow the Find Object task, wherein each agent is given multiple attempts to find a target object. If the agent correctly predicts the furniture object on which the object is placed, we count the event as a success and the task is terminated. In this setup, given a target object, the agent makes $N$ predictions of furniture objects, which are sequentially visited by the agent. The cost returned by the environment is therefore the total path length traversed by the agent in an episode, and the number of attempts it took to find the target object ($N+1$ if object could not be found in $N$ steps). It should be noted that the path length itself is not a direct indicator of performance, as the furniture node on which an object is placed can sometimes be far away from the agent's current position. Hence, the optimal solution in this case would incur a high path length. As a result, even optimal agents which can find the target object in a few steps can have a high variance for the average path length when aggregated over thousands of episodes.

\begin{table}[!ht]
    \centering
    \caption{\label{tab:sampleable_furniture} Furniture types used in Dynamic House Sim and their associated metadata } 
    \resizebox{0.5\textwidth}{!}{%
    \begin{tabular}{|l|l|l|l|l|}
    \hline
        \textbf{label} & \textbf{\# possible adjectives} & \textbf{sample probability} & \textbf{max count} & \textbf{\# edges} \\ \hline
        counter & 3 & 0.80 & 3 & 27 \\ \hline
        table & 1 & 0.80 & 3 & 17 \\ \hline
        shelf & 7 & 0.80 & 10 & 6 \\ \hline
        fridge & 3 & 0.90 & 2 & 30 \\ \hline
        top cabinet & 3 & 0.50 & 10 & 8 \\ \hline
        coffee table & 4 & 0.60 & 3 & 34 \\ \hline
        cooktop & 3 & 0.90 & 1 & 3 \\ \hline
        counter top & 5 & 0.80 & 3 & 67 \\ \hline
        dining table & 7 & 1.00 & 1 & 74 \\ \hline
        chair & 7 & 0.75 & 12 & 40 \\ \hline
        tv stand & 2 & 1.00 & 1 & 33 \\ \hline
        sofa & 1 & 0.75 & 2 & 18 \\ \hline
        bed & 3 & 0.90 & 2 & 17 \\ \hline
        dresser & 7 & 0.75 & 3 & 36 \\ \hline
        toilet & 1 & 1.00 & 1 & 15 \\ \hline
        sink & 2 & 1.00 & 2 & 6 \\ \hline
        shelving unit & 4 & 0.50 & 4 & 28 \\ \hline
        desk & 7 & 0.70 & 4 & 30 \\ \hline
        chair & 4 & 0.60 & 2 & 17 \\ \hline
        side table & 7 & 0.70 & 3 & 64 \\ \hline
        chairs & 7 & 0.75 & 12 & 40 \\ \hline
        couch & 1 & 0.75 & 2 & 18 \\ \hline
    \end{tabular}
    }
\end{table}

\pagebreak
{\scriptsize
\begin{longtable}[!h]{|l|l|l|l|l|l|l|}
    \caption{Objects types used in Dynamic House Sim and their associated metadata}
    \label{tab:sampleable_objs}\\
    \hline
        \textbf{label} & \textbf{\# adjectives} & \textbf{sample prob} & \textbf{max count} & \textbf{move frequency} & \textbf{add/remove prob} & \textbf{\# edges} \\ \hline
        apple & 2 & 0.80 & 15 & 0.40 & 0.20 & 9 \\ \hline
        box & 6 & 0.70 & 4 & 0.40 & 0.01 & 11 \\ \hline
        cereal & 9 & 0.50 & 4 & 0.20 & 0.10 & 2 \\ \hline
        dishtowel & 11 & 0.75 & 8 & 0.20 & 0.00 & 1 \\ \hline
        flour & 8 & 0.80 & 4 & 0.20 & 0.10 & 2 \\ \hline
        jar & 10 & 0.75 & 6 & 0.15 & 0.05 & 3 \\ \hline
        kettle & 6 & 0.75 & 4 & 0.40 & 0.00 & 4 \\ \hline
        lettuce & 1 & 0.80 & 5 & 0.10 & 0.01 & 3 \\ \hline
        milk & 6 & 0.80 & 2 & 0.10 & 0.10 & 2 \\ \hline
        mug & 10 & 0.80 & 12 & 0.60 & 0.01 & 11 \\ \hline
        oil & 9 & 0.80 & 4 & 0.10 & 0.05 & 2 \\ \hline
        pasta & 6 & 0.75 & 8 & 0.10 & 0.00 & 2 \\ \hline
        rice & 11 & 0.20 & 6 & 0.15 & 0.00 & 2 \\ \hline
        soda & 12 & 0.50 & 8 & 0.10 & 0.10 & 2 \\ \hline
        ladle & 9 & 0.80 & 6 & 0.40 & 0.00 & 1 \\ \hline
        toy & 3 & 0.80 & 12 & 0.50 & 0.01 & 0 \\ \hline
        egg & 2 & 0.80 & 12 & 0.05 & 0.10 & 2 \\ \hline
        spray bottle & 6 & 0.50 & 4 & 0.33 & 0.00 & 6 \\ \hline
        salt shaker & 4 & 0.50 & 2 & 0.25 & 0.00 & 3 \\ \hline
        wine bottle & 7 & 0.40 & 6 & 0.10 & 0.10 & 3 \\ \hline
        potato & 1 & 0.75 & 8 & 0.10 & 0.00 & 3 \\ \hline
        pencil & 4 & 0.50 & 12 & 0.50 & 0.01 & 11 \\ \hline
        soap bottle & 7 & 0.50 & 4 & 0.10 & 0.01 & 4 \\ \hline
        plate & 7 & 0.80 & 12 & 0.40 & 0.00 & 9 \\ \hline
        fork & 6 & 0.80 & 8 & 0.20 & 0.20 & 3 \\ \hline
        book & 7 & 0.80 & 20 & 0.20 & 0.01 & 14 \\ \hline
        pan & 9 & 0.75 & 6 & 0.20 & 0.05 & 2 \\ \hline
        towel roll & 2 & 0.75 & 4 & 0.25 & 0.00 & 3 \\ \hline
        butter knife & 2 & 0.75 & 4 & 0.20 & 0.01 & 3 \\ \hline
        spoon & 9 & 0.75 & 16 & 0.20 & 0.00 & 2 \\ \hline
        watch & 4 & 0.50 & 2 & 0.40 & 0.00 & 8 \\ \hline
        phone & 7 & 0.75 & 2 & 0.90 & 0.00 & 13 \\ \hline
        pen & 6 & 0.75 & 8 & 0.50 & 0.00 & 9 \\ \hline
        credit card & 4 & 0.50 & 4 & 0.20 & 0.10 & 9 \\ \hline
        candle & 8 & 0.60 & 12 & 0.40 & 0.20 & 6 \\ \hline
        tissue box & 6 & 0.20 & 4 & 0.10 & 0.10 & 5 \\ \hline
        newspaper & 8 & 0.60 & 5 & 0.60 & 0.20 & 12 \\ \hline
        remote control & 4 & 0.75 & 4 & 0.75 & 0.00 & 12 \\ \hline
        house plant & 8 & 0.75 & 12 & 0.01 & 0.01 & 8 \\ \hline
        laptop & 13 & 0.60 & 4 & 0.75 & 0.10 & 12 \\ \hline
        desk lamp & 4 & 0.50 & 4 & 0.01 & 0.01 & 7 \\ \hline
        alarm clock & 4 & 0.20 & 1 & 0.01 & 0.00 & 13 \\ \hline
        soap bar & 7 & 0.50 & 4 & 0.15 & 0.01 & 2 \\ \hline
        toilet paper & 1 & 0.50 & 4 & 0.10 & 0.00 & 1 \\ \hline
        baseball bat & 2 & 0.20 & 1 & 0.20 & 0.00 & 8 \\ \hline
        dish sponge & 7 & 0.80 & 4 & 0.10 & 0.01 & 4 \\ \hline
        tennis racket & 1 & 0.25 & 4 & 0.20 & 0.00 & 5 \\ \hline
        basket ball & 1 & 0.20 & 1 & 0.20 & 0.00 & 11 \\ \hline
        coffee machine & 3 & 0.60 & 2 & 0.01 & 0.00 & 2 \\ \hline
        knife & 1 & 0.60 & 12 & 0.20 & 0.01 & 2 \\ \hline
        bread & 4 & 0.50 & 4 & 0.14 & 0.10 & 2 \\ \hline
        cup & 11 & 0.80 & 16 & 0.40 & 0.10 & 3 \\ \hline
        pot & 9 & 0.50 & 4 & 0.10 & 0.00 & 4 \\ \hline
        bottle & 11 & 0.90 & 15 & 0.25 & 0.10 & 5 \\ \hline
        toaster & 2 & 0.80 & 1 & 0.01 & 0.00 & 2 \\ \hline
        cloth & 8 & 0.90 & 6 & 0.20 & 0.01 & 2 \\ \hline
        microwave & 2 & 0.80 & 1 & 0.00 & 0.00 & 2 \\ \hline
        apples & 2 & 0.80 & 15 & 0.40 & 0.20 & 9 \\ \hline
        oranges & 2 & 0.80 & 15 & 0.40 & 0.20 & 9 \\ \hline
        bananas & 2 & 0.80 & 15 & 0.40 & 0.20 & 9 \\ \hline
        orange & 2 & 0.80 & 15 & 0.40 & 0.20 & 9 \\ \hline
        banana & 2 & 0.80 & 15 & 0.40 & 0.20 & 9 \\ \hline
        lemon & 2 & 0.80 & 15 & 0.40 & 0.20 & 9 \\ \hline
        garlic & 2 & 0.80 & 15 & 0.40 & 0.20 & 9 \\ \hline
        peach & 2 & 0.80 & 15 & 0.40 & 0.20 & 9 \\ \hline
        grapes & 2 & 0.80 & 15 & 0.40 & 0.20 & 9 \\ \hline
        avocado & 2 & 0.80 & 15 & 0.40 & 0.20 & 9 \\ \hline
        towels & 11 & 0.75 & 8 & 0.20 & 0.00 & 1 \\ \hline
        beet & 1 & 0.80 & 5 & 0.10 & 0.01 & 3 \\ \hline
        radish & 1 & 0.80 & 5 & 0.10 & 0.01 & 3 \\ \hline
        eggplant & 1 & 0.80 & 5 & 0.10 & 0.01 & 3 \\ \hline
        basil & 1 & 0.80 & 5 & 0.10 & 0.01 & 3 \\ \hline
        tomato & 1 & 0.80 & 5 & 0.10 & 0.01 & 3 \\ \hline
        kale & 1 & 0.80 & 5 & 0.10 & 0.01 & 3 \\ \hline
        squash & 1 & 0.80 & 5 & 0.10 & 0.01 & 3 \\ \hline
        yogurt & 6 & 0.80 & 2 & 0.10 & 0.10 & 2 \\ \hline
        whole fat milk & 6 & 0.80 & 2 & 0.10 & 0.10 & 2 \\ \hline
        zero fat milk & 6 & 0.80 & 2 & 0.10 & 0.10 & 2 \\ \hline
        pop & 12 & 0.50 & 8 & 0.10 & 0.10 & 2 \\ \hline
        teddy bear & 3 & 0.80 & 12 & 0.50 & 0.01 & 0 \\ \hline
        legos & 3 & 0.80 & 12 & 0.50 & 0.01 & 0 \\ \hline
        action figure & 3 & 0.80 & 12 & 0.50 & 0.01 & 0 \\ \hline
        dinosaur & 3 & 0.80 & 12 & 0.50 & 0.01 & 0 \\ \hline
        jigsaw & 3 & 0.80 & 12 & 0.50 & 0.01 & 0 \\ \hline
        animal & 3 & 0.80 & 12 & 0.50 & 0.01 & 0 \\ \hline
        butter & 2 & 0.80 & 12 & 0.05 & 0.10 & 2 \\ \hline
        pepper shaker & 4 & 0.50 & 2 & 0.25 & 0.00 & 3 \\ \hline
        paprika shaker & 4 & 0.50 & 2 & 0.25 & 0.00 & 3 \\ \hline
        bottle of soap & 7 & 0.50 & 4 & 0.10 & 0.01 & 4 \\ \hline
        plates & 7 & 0.80 & 12 & 0.40 & 0.00 & 9 \\ \hline
        binder & 7 & 0.80 & 20 & 0.20 & 0.01 & 14 \\ \hline
        document & 7 & 0.80 & 20 & 0.20 & 0.01 & 14 \\ \hline
        books & 7 & 0.80 & 20 & 0.20 & 0.01 & 14 \\ \hline
        binders & 7 & 0.80 & 20 & 0.20 & 0.01 & 14 \\ \hline
        documents & 7 & 0.80 & 20 & 0.20 & 0.01 & 14 \\ \hline
        spoons & 9 & 0.75 & 16 & 0.20 & 0.00 & 2 \\ \hline
        smartphone & 7 & 0.75 & 2 & 0.90 & 0.00 & 13 \\ \hline
        wallet & 4 & 0.50 & 4 & 0.20 & 0.10 & 9 \\ \hline
        debit card & 4 & 0.50 & 4 & 0.20 & 0.10 & 9 \\ \hline
        candles & 8 & 0.60 & 12 & 0.40 & 0.20 & 6 \\ \hline
        box of tissues & 6 & 0.20 & 4 & 0.10 & 0.10 & 5 \\ \hline
        pc & 13 & 0.60 & 4 & 0.75 & 0.10 & 12 \\ \hline
        bar of soap & 7 & 0.50 & 4 & 0.15 & 0.01 & 2 \\ \hline
        soap & 7 & 0.50 & 4 & 0.15 & 0.01 & 2 \\ \hline
        dish soap & 7 & 0.80 & 4 & 0.10 & 0.01 & 4 \\ \hline
        sponge & 7 & 0.80 & 4 & 0.10 & 0.01 & 4 \\ \hline
        baguette & 4 & 0.50 & 4 & 0.14 & 0.10 & 2 \\ \hline
        bottles & 11 & 0.90 & 15 & 0.25 & 0.10 & 5 \\ \hline
\end{longtable}
}